\documentclass{article}

\usepackage{graphicx} 
\usepackage{subfigure} 
\usepackage{natbib}
\usepackage{algorithm}
\usepackage{algorithmic}
\usepackage{hyperref}
\usepackage{multirow}
\usepackage{verbatim}

\usepackage[accepted]{icml2013} 

\icmltitlerunning{Song-based Classification techniques for Endangered Bird Conservation}

\begin{document} 

\twocolumn[
\icmltitle{Song-based Classification techniques for Endangered Bird Conservation}

\icmlauthor{Erick Stattner}{estattne@univ-ag.fr}
\icmladdress{LAMIA Lab. University of the French West Indies and Guiana, France}
            
\icmlauthor{Wilfried Segretier}{wsegreti@univ-ag.fr}
\icmladdress{LAMIA Lab. University of the French West Indies and Guiana, France}

\icmlauthor{Martine Collard}{mcollard@univ-ag.fr}
\icmladdress{LAMIA Lab. University of the French West Indies and Guiana, France}

\icmlauthor{Philippe Hunel}{phunel@univ-ag.fr}
\icmladdress{LAMIA Lab. University of the French West Indies and Guiana, France}

\icmlauthor{Nicolas Vidot}{nvidot@martinique.univ-ag.fr}
\icmladdress{LAMIA Lab. University of the French West Indies and Guiana, France}

\icmlkeywords{acoustical data, machine learning, automatic observation}

\vskip 0.3in
]

\begin{abstract} 
	The work presented in this paper is part of a global framework which long term goal is to design a wireless sensor network able to support the observation of a population of endangered birds. We present the first stage for which we have conducted a knowledge discovery approach on a sample of acoustical data. 
We use MFCC features extracted from bird songs and we exploit two knowledge discovery techniques.
One relies on clustering-based approaches and highlights the homogeneity in the songs of the species.
The other one is based on predictive modeling and demonstrates the good performances of various machine learning techniques for the identification process.
The knowledge elicited provides promising results to consider a widespread study and to elicit guidelines for designing a first version of the automatic approach for data collection based on acoustic sensors.

\end{abstract}

\section{Introduction}
In last decades, due to the exponential growth of global commercial and industrial activities, numerous scientists have focused on environmental and social troubles that are becoming increasingly worrying: global warming, pollution, disease spreading, exhaustion of energy resources or biodiversity assessment.

This last phenomenon is one of the most challenging problem for geologists, ecologists, biologists and ethologists.
Indeed, it is now apparent that changes occurring on an environment, as small as they are, can have significant impacts on the equilibrium of an ecosystem, and especially on the survival of animal species that depend on it.

New information and communication technologies and devices provide powerful tools for collecting useful and wide scale information on a variety of factors potentially involved in biodiversity loss. For example, GPS devices have been used for tracking individual movements in situations in which a human presence is not possible~\cite{Rumble2001,Ryan2004}. Fixed devices such as sensors are considered for detecting the presence of individuals ~\cite{Fagerlund2007,Cai2007} and studying their behavior~\cite{Stattner2010-MoMM, Stattner2011-LCN}.

\begin{figure}
	\center
	\includegraphics[scale=0.10]{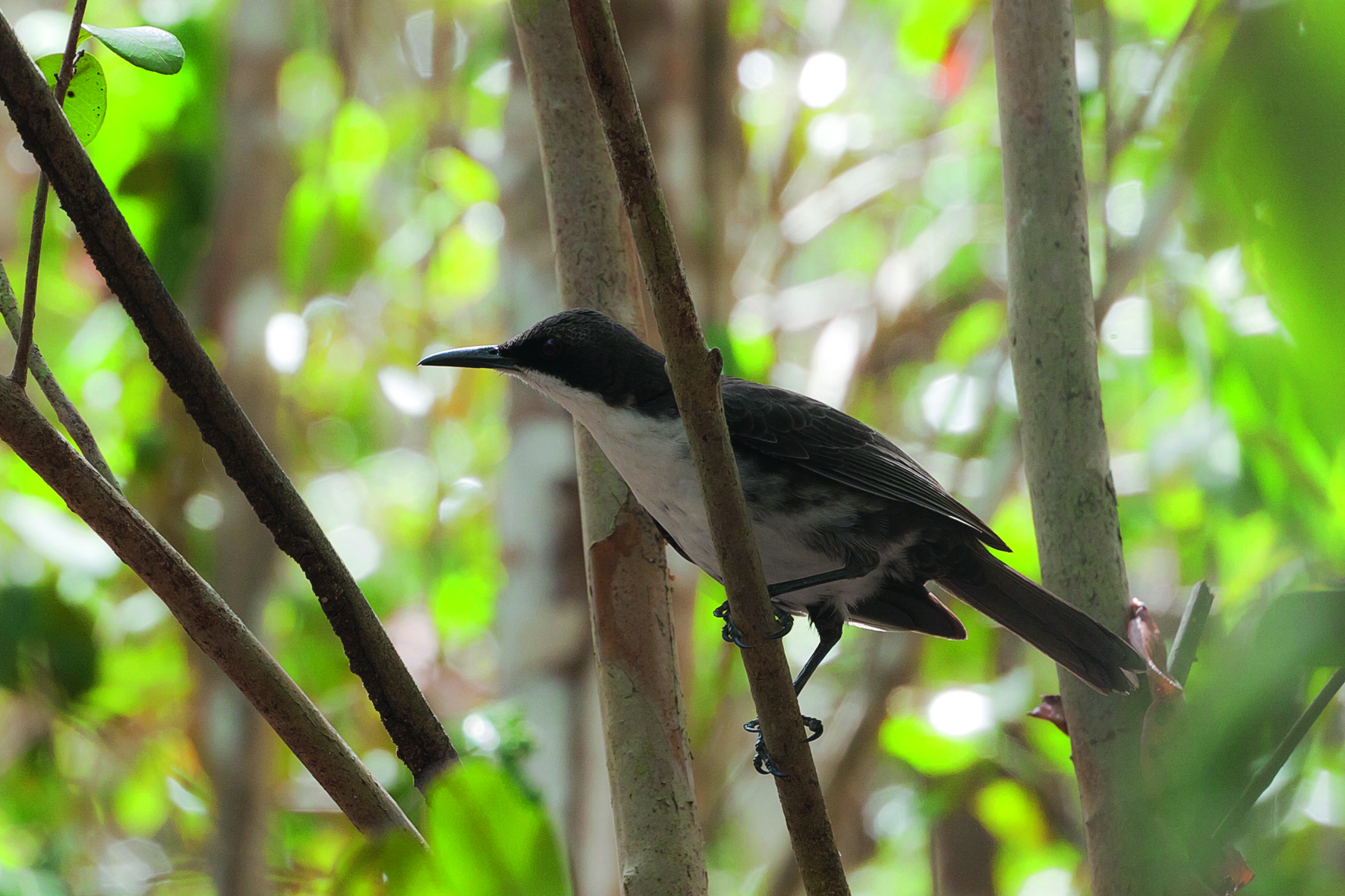}
	\caption{An individual of Moqueur Gorge Blanche in its habitat (\copyright Vincent Lemoine)}
	\label{figMGB}
\end{figure}

In this paper, a case study is presented on an endangered bird species endemic to the site of \textit{La Caravelle} on the Martinique Island, French West Indies, called the \textit{Moqueur Gorge Blanche} (MGB), for which local scientists have initiated a work of both study and protection (see Figure~\ref{figMGB}).
Since it is quite tedious and unproductive to manually collect data by visual observation during programs conducted on the natural area of the species, the final goal is to design an efficient methodology to automate the collection. Indeed although the manual technique has helped to obtain a first view of the species behavior, it also raises many questions about the reliability of the data collected, since 
(i)~data are collected only during short periods,
(ii)~the human presence may, itself, affect the bird behavior.

Our long-term objective is to design an automatic solution for data collection, based on a wireless acoustic sensor network, able to capture widespread information about the bird population. Each sensor of the network should be fitted with a microphone and be able to detect the presence of the species by analysing the songs.

In this paper, we address the latter issue to evaluate how song analysis is relevant for predicting the presence of the species with a good precision on the basis of their songs. In this preliminary step, we have considered a set of pre-recorded songs of the Moqueur Gorge Blanche~\cite{CD_MGB}. We have trained different knowledge discovery techniques on the corresponding signals and we have obtained encouraging results to consider an efficient knowledge extraction from acoustical data and an optimal design of the sensor network.

The papers is organized as follows.
Section~\ref{secRelated} presents standard recognition processes on audio signals. Section~\ref{secAcoustic} is devoted to the description of our case study data and to their pre-processing. In Section~\ref{secAutomatic} we discuss the knowledge discovery approach conducted on pre-processed data. In Section~\ref{secConclusion} after a short synthesis, we present further works to lead on the project.

\section{Related works} \label{secRelated}
Last years, the analysis of acoustic signals has been a very active research area that have found applications in various domain such as speech recognition~\cite{Sakoe1978}, speaker identification~\cite{Reynolds1995}, source localization~\cite{Valin2003}.
Globally, the recognition process performs in two key steps.

\textbf{(i)~A parametrization process}, that summarizes the recorded audio signal through a characteristic fingerprint.
This fingerprint is designed so that if two records are similar, their associated fingerprints should also be very close.
More specifically, the parametrization process allows representing the audio signals by a series of coefficients that describe it~\cite{Eisele1996}.
Several parametrization techniques have been proposed such as LPCC (Linear Prediction Cepstral Coefficients) or MFCC (Mel-Frequency Cepstral Coefficients).
As our final objective is to implement our recognition process on sensors of a network, we use in this work the MFCC  (Mel Frequency Cepstral Coefficient) technique, since it has been shown that this technique has good performances on this kink of devices~\cite{Levy2003}.

\textbf{(ii)~A classification process}, that aims to determine if the generated fingerprint belongs to a known class~\cite{Rabiner1979}.
For instance, in the particular case of the recognition of bird songs, Fagerlund~\cite{Fagerlund2007} uses a support vector machine to identify songs of a given bird species, while Cai et al.~\cite{Cai2007} suggest the use of a neural network.
\\
As our objective is to perform the recognition on sensors, that are known to have limited capacities in terms of calculation and memory, we use in this work standard classification algorithm with the objective to obtain a predictive model easy to implements on the motes of the network.

\section{Acoustical data} \label{secAcoustic}
In this section, we explain how original audio raw data have been preprocessed to produce acoustical datasets on which knowledge discovery techniques have been applied.

The initial audio data or \textit{raw data} are songs of Moqueur Gorge Blanche collected in 2009 on the site of La Caravelle~\cite{CD_MGB}. 
We have extracted seven song examples.
Each song is sampled at 44.1 kHz and stored as 16-bit signed mono .wav format files.

\begin{figure}[!h]
	\centering
	\subfigure[]{
		\includegraphics[width=6cm,height=5cm]{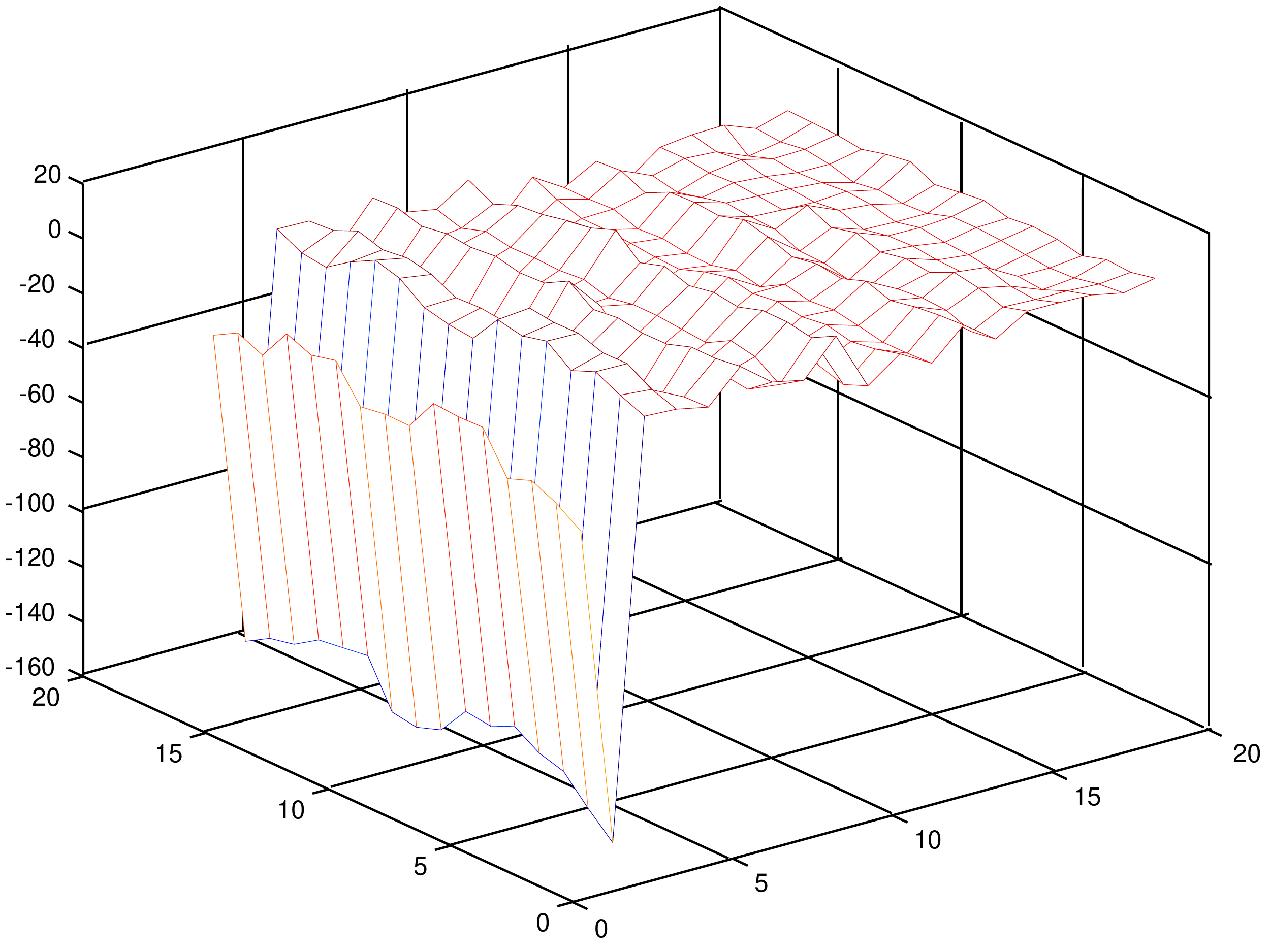}
	}
	\subfigure[]{
		\includegraphics[width=5.5cm,height=3.5cm]{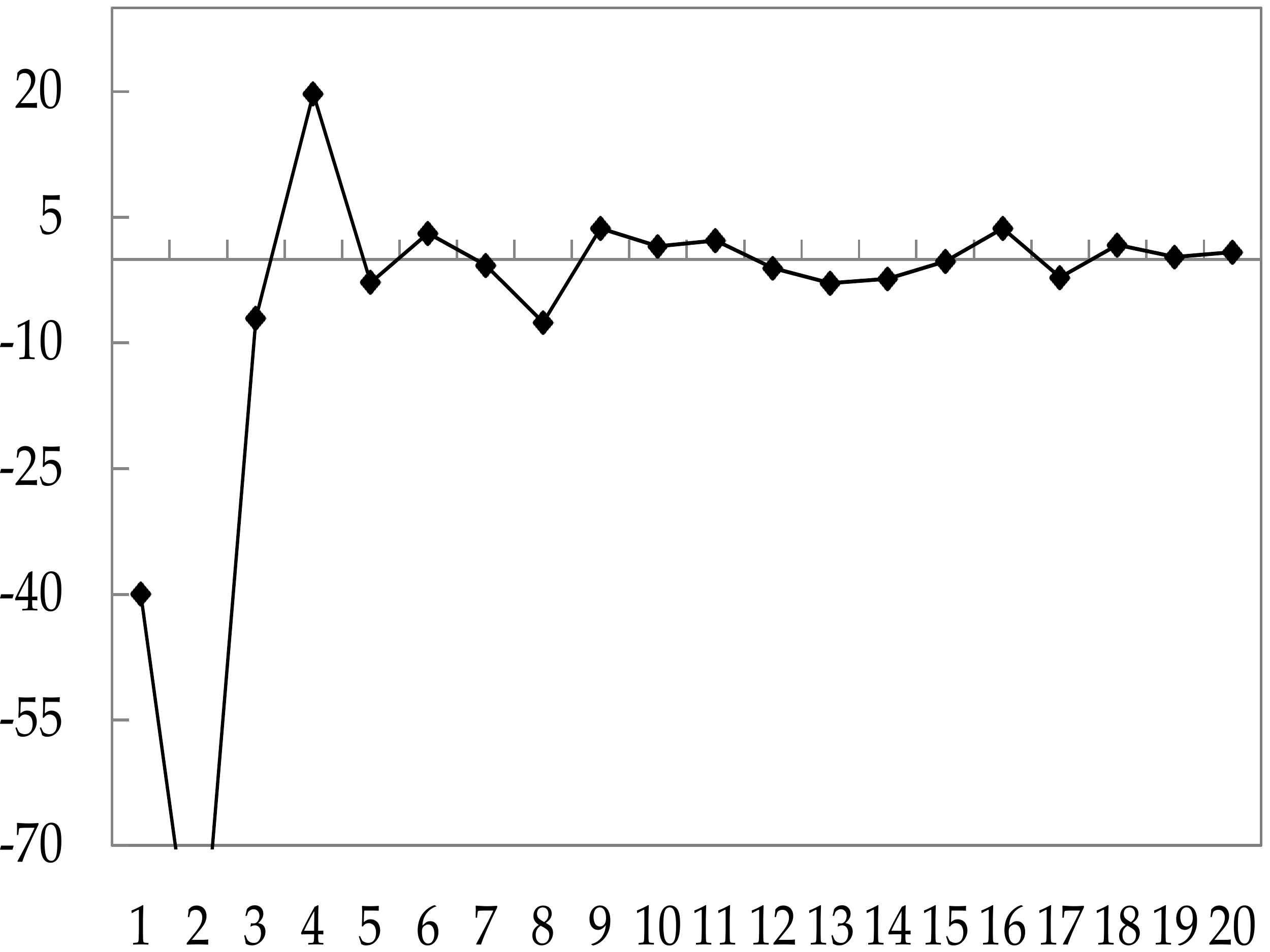}
	}
	\caption{Example of (a)~MFCC and (b)~average fingerprint, obtained from a song of the Moqueur Gorge Blanche}
	\label{figMfcc}
\end{figure}

For each song example, MFCC features have been computed with the Java framework CoMIRVA~\cite{Schedl2007} developed and maintained by Markus Schedl (see Figure~\ref{figMfcc}(a)).
In order to simplify the computation phase, we have generated a mean fingerprint for each song, by averaging the different MFCC values provided for each coefficient (see Figure~\ref{figMfcc}(b)). Thus we have a mean fingerprint vector of 20 averaged MFCC coefficients for each of the 7 songs. Each of the 20 averaged MFCC coefficients corresponds to a time window of an original song.
\\
In this way, the problem is to determine, for such a fingerprint, whether it belongs to the species or not. In other words, we would like to predict with a good accuracy that a new occurring song belongs to the Moqueur Gorge Blanche.

In order to conduct classification tasks, we have complemented these examples with counter-examples to obtain a training dataset of mean fingerprints obtained on other bird species.
Thus, 17 new examples have been added for species such as Common blackbird, Accentor, Heron, Woodcock, etc. 

Traditionally the data mining \textit{preprocessing} step is very important since choices have to be done for preparing, cleaning and transforming the raw data that may be often noisy, imbalanced and not in the adequate format that learning algorithms requires as input. In the current study, as described just above, the original audio data are transformed to produce the dataset composed of 24 averaged MFCC fingerprints.\\
For predictive modelling, it is necessary to label each data sample with a corresponding class name. In this work we address a typical binary classification problem to predict if a data sample is an actual Moqueur Gorge Blanche song ("MGB" class) or another bird species song ("Other" class). In the following, the "MGB" class will be considered as the positive class while the "Other" class will be considered as negative. As said before, the "MGB" class contains 7 examples while the "Other" contains 17 examples. Our dataset is thus clearly imbalanced since there are more than twice as much "Other" examples than "MGB". 
In order to overcome the imbalance between classes in datasets, two strategies are commonly used:
\begin{enumerate}
\item assign distinct costs to class examples (usually higher costs for the
  minority class) \cite{PazzaniMMAHB94}
\item re-sample the source dataset, either by over-sampling the minority class and/or under-sampling the majority class \cite{KubatM97}
\end{enumerate}
Since the dataset is rather short, we have followed the second approach to create balanced training sets. We have thus over-sampled the "MGB" class using the {\it Synthetic Minority Over-sampling Technique} (SMOTE) algorithm \cite{ChawlaBHK02} which consists in generating synthetic examples by randomly selecting points along the lines that join a minority class original sample and some of its nearest neighbors. Since this over-sampling technique is based on random properties, we have generated 100 extended balanced datasets (with 16 "MGB" and 17 "Other") and we have averaged the performances obtained on each of them in order to have statistically significant results.

In the following, we have conducted experiments on both datasets without and with over-sampling. In the following, we refer to them as Simple\_MFCC and Extended\_MFCC dataset respectively.

\section{Classification-based automatic detection} \label{secAutomatic}

We describe in this section the knowledge discovery method applied that covers both descriptive and predictive approaches. On the \textit{descriptive} axis, we have conducted a clustering based on dynamic time warping (DTW), a time series alignment algorithm developed originally for the purpose of speech recognition~\cite{Sakoe1978}, and on the \textit{predictive} axis, we have trained classical algorithms. On the one hand, the clusters extracted that have a very good matching with the classes "MGB" and "Other", tend to show the homogeneity of MGB song signals. On the other hand, \textit{predictions} obtained by training on the Simple\_MFCC and Extended\_MFCC datasets provide good performances that allow us to consider further wide scale training and testing experiments.

\subsection{Clustering} \label{subClustering}
In the area of the analysis of acoustic signals, the underlying assumption is that it exists a high degree of similarity between the elements of a same class.
In other words, we can expect that the examples of a same class also belong to a same cluster.

Thus, in a first approach, we have studied the matching between the clusters and the classes.
The matching is based on the notion of distance between the examples.
For evaluating the distance between two examples, we use the classical DTW algorithm commonly use in the area of signal processing.
\\
More precisely, let $s_{1}$ and $s_{2}$ be two sequences, the objective of DTW is to align these sequences by warping the time axis iteratively until an optimal match is found.
The originality of this algorithm is its capacity to evaluate the similarity between two sequences that may vary in size, time or speed.

Our approach was as follows:
\\
(i)~A reference fingerprint has been created by averaging all fingerprints of examples of Moqueur Gorge Blanche.
\\
(ii)~For a given song fingerprint, we measure, by using the DTW algorithm, the distance to the reference fingerprint that characterizes the species.
When this distance is under a given threshold $\beta$, the fingerprint is closed to the reference and we suppose that the associated song belongs to the species. 
Otherwise, the song is identified as different from that species.

\begin{figure}[!h]
	\centering
	\subfigure[]{
		\includegraphics[width=6cm,height=4cm]{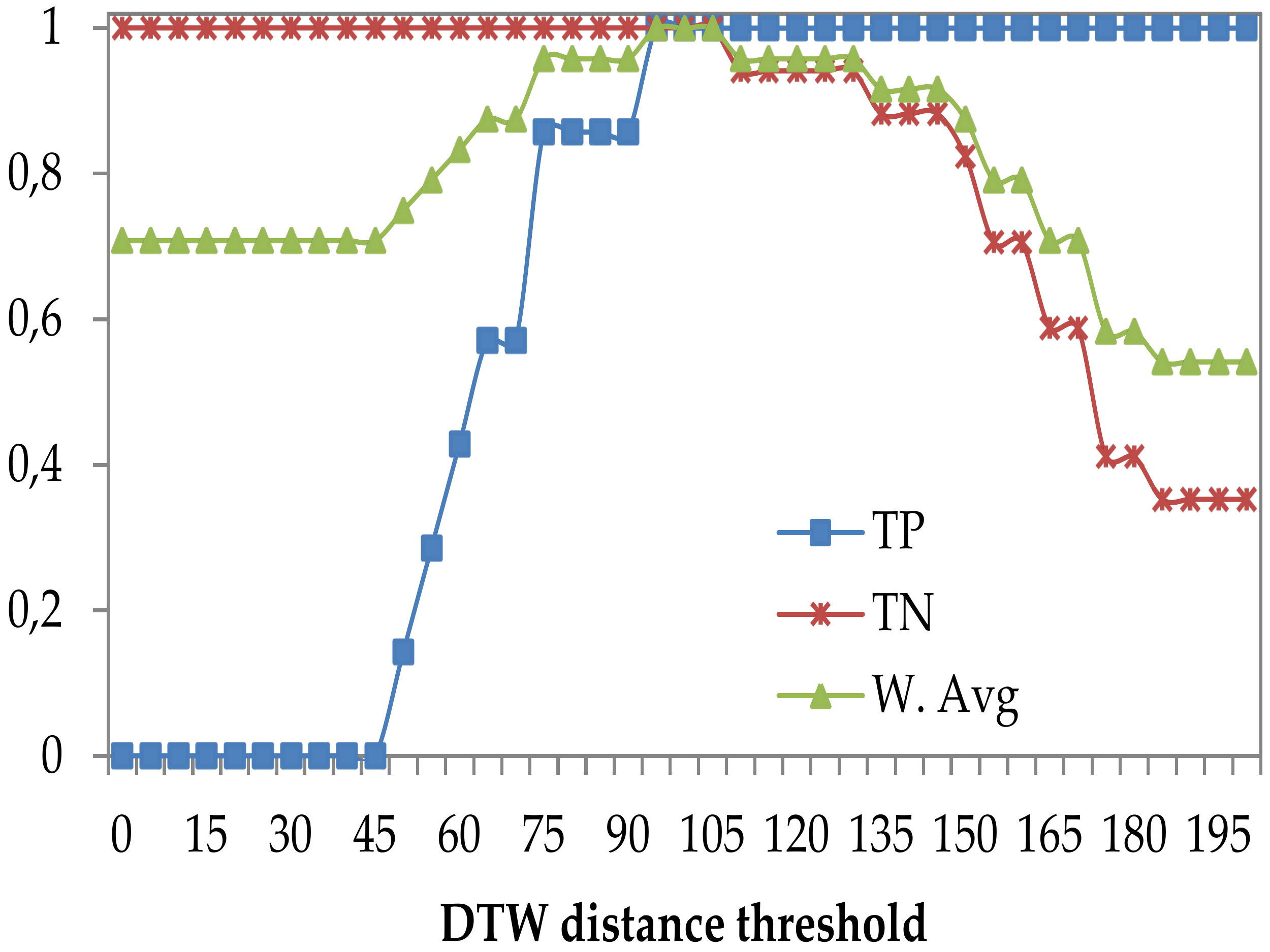}
	}
	\subfigure[]{
		\includegraphics[width=6cm,height=4cm]{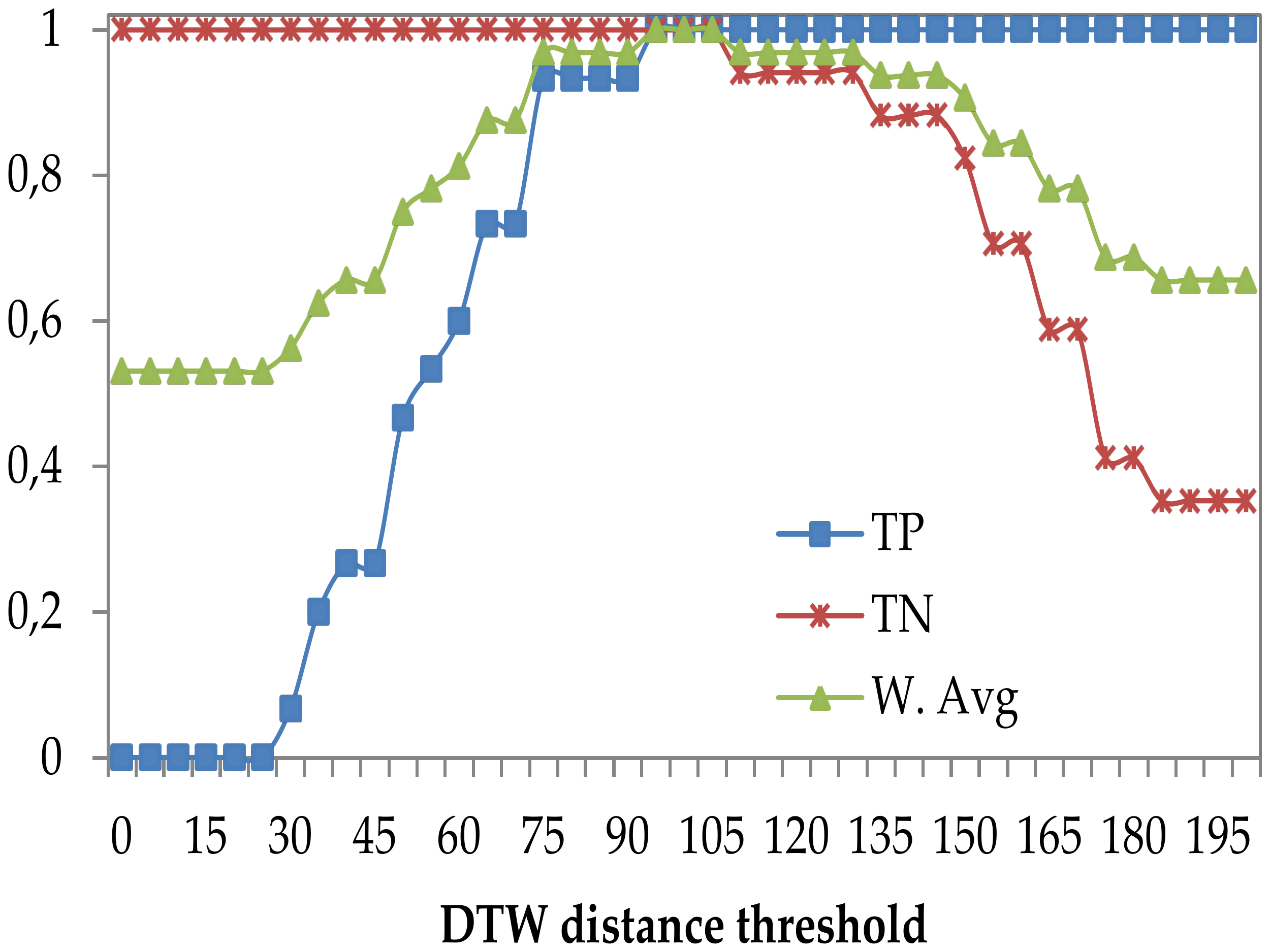}
	}
	\caption{Performances on the identification process based on DTW distance with (a)~simple (b)~extended datasets (TP: True positive, TN: True negative, $W.~Avg$ performances)}
	\label{figRsltCluster}
\end{figure}

By using both datasets (simple and extended), we have measured the performances of the identification when using the distance with DTW.
Figure~\ref{figRsltCluster} shows these results according to the threshold $\beta$.
In this clustering context, for a given cluster, we call TP, or True Positive rate, the fraction of "MGB" class examples into this cluster for which the DTW distance is under the $\beta$ threshold. Similarly, we call TN, or True Negative rate, the fraction of "Other" class examples into the cluster for which the DTW distance is above the $\beta$ threshold.
$W.~Avg$ gives information about the performances of the detection by measuring the following ratio $\frac{|MGB| \times TP + |Other| \times TN}{|MGB|+|Other|}$.

When the DTW distance threshold $\beta$ is very low, all examples are detected as belonging to the ``Other'' cluster. This explains why $TN$ is very high and $TP$ very low.
Inversely, when $\beta$ is high, a lot of examples are detected as belonging to the $MGB$ cluster, which explains the high degree of $TP$ and the low degree of $TN$.
\\
However, we can observe that it exists a DTW distance threshold for which the detection is maximal.
Indeed, when $\beta \in [95..110]$, we observe that $TP$, $TN$ and $W.~Avg$ are all equal to $1$.
These results suggest that the MGB songs tend to be very homogeneous, since we can find some distance thresholds for which all examples of a same class belong to the same cluster.

\subsection{Predictive modelling} \label{subPredictive}
\begin{table*}[!ht]
    \caption{Performances of standard machine learning algorithms averaged on 100 leave one out cross validation experiments}
    \begin{center}
\begin{small}
\begin{tabular}{|c c|c c c|c c c| }
\hline
 Technique&&\multicolumn{3}{c|}{Simple\_MFCC}&\multicolumn{3}{c|}{Extented\_MFCC}\\
& &TN&TP&W.Avg&TN&TP&W.Avg\\ \hline

\multirow{2}{*}{C4.5}&Avg&88.20&71.40&83.33&91.66&88.2&89.88\\
&Std Dev&-&-&-&2.98&0.00&1.42\\
\multirow{2}{*}{RF}&Avg&86.99&90.33&89.58&98.32&90.22&94.14\\
&Std Dev&9.29&5.06&4.23&2.89&5.39&2.99\\
\multirow{2}{*}{NB}&Avg&88.20&85.7&87.5&96.21&100.00&98.17\\
&Std Dev&-&-&-&3.02&0.00&1.46\\
\multirow{2}{*}{MLP}&Avg&82.51&100&87.58&100&83.84&91.65\\
&Std Dev&0.00&0.81&0.59&0.00&3.12&1.61\\
\hline
\end{tabular}
\end{small}
    \end{center}
    \label{Results_classif}
\end{table*}

In this section we present the results that we obtained in the area of supervised machine learning with the following commonly used machine learning techniques: C4.5 \cite{quinlan1993} and Random Forest (RF) \cite{Breiman2001} decision tree approaches, Naive Bayes (NB) \cite{Rish2001} and Multi-layer Perceptron (MLP) Artificial Neural Network (ANN). Since the total number of examples was quite low in order to generate typical 66\%/33\% train and test datasets, we used a leave-one-out cross validation (loocv) scheme (i.e. k-fold cross validation with k set as the total number of available examples minus one) to estimate the performances of each experiment conducted in this section.

\begin{figure}[!h]
\centering
   \includegraphics[width=8cm]{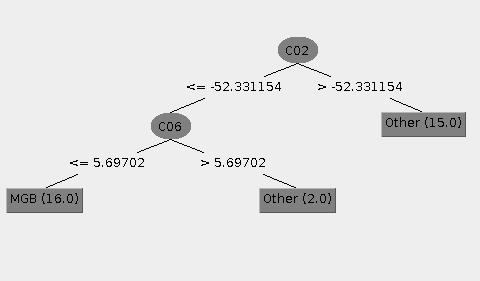}
\caption{Example of decision tree obtained with C4.5.}  
\label{J48} 
\end{figure}

Table \ref{Results_classif} presents the performances obtained by each technique with and without the synthetic examples generated by the SMOTE algorithm. First column refers to the technique used, while columns TN, TP and W.Avg stands for True Negative, True Positive and Weighted Average ($\frac{|MGB| \times TP + |Other| \times TN}{|MGB|+|Other|}$) rates. In this context of supervised classification, true positive rates correspond, as usually, to the percentages of "MGB" examples correctly classified as "MGB" while true negative rates stand for the percentages of "Other" examples correctly classified as "Other". Since each algorithm has been launched 100 times to overcome either its own random side and/or the randomness introduced by SMOTE, we give averaged performances rates and their associated standard deviation as a confidence\footnote{Only deterministic techniques C4.5 and Naive Bayes have been ran (with loocv) once on the original data and thus do not have std dev values.}.\\
We can see that for the Simple\_MFCC dataset, the best weighed average performances are obtained with Random Forest (89.8\%) while for the extented datasets best results are given by Naive Bayes (98.52\%). Even if no clear tendency can be observed concerning the repartition of TP and TN rates in both types of datasets, the overall performances obtained on extented data seem to be significantly better.
Figure \ref{J48} shows an example of decision tree learned with the C4.5 algorithm. The leaves show the final classification decisions taken according to the values observed on selected attributes (represented by the nodes). For instance, in this case, if the value observed on the attribute "C02" is lower or equal than -52.33 and the value observed on "C06" is greater than 5.69, the "Other" class is predicted.

\section{Conclusion and future directions} \label{secConclusion}
Our overall project is to design a wireless sensor network able to support the observation of an endangered birds species endemic to the Martinique island, called the Moqueur Gorge Blanche.
\\
This paper has focused on the first stage of the work, by evaluating how song analysis is relevant for predicting the presence the species with a good precision.

For this purpose, we have shown how knowledge discovery techniques can be used for recognize the songs of the species.
The results obtained, that highlight the good performances of these techniques for the recognition of the Moqueur Gorge Blanche songs, allow to consider a real deployment on the ground

In our future works, our objective is to implement the recognition process on a wireless sensor network, in which sensors are fitted with microphone.
In a first step, we plan to use the data collected on given periods for optimizing the network configuration.

A first track should be to use the sensor to identify the regions in which the population density is high. More sensors could thus be positioned in such regions, while less sensors would be allocated to the regions less frequented by the species.

At long terms, the data collected on the presence of the species could be used to search for correlations between the features of the habitat and the presence of individuals.
Such a knowledge could have very relevant applications for the preservation of the species.
For instance, it could be used for recreating a favorable habitat for the species.

\bibliographystyle{icml2013}
\bibliography{biblioICML2013}

\begin{thebibliography}{20}
\providecommand{\natexlab}[1]{#1}
\providecommand{\url}[1]{\texttt{#1}}
\expandafter\ifx\csname urlstyle\endcsname\relax
  \providecommand{\doi}[1]{doi: #1}\else
  \providecommand{\doi}{doi: \begingroup \urlstyle{rm}\Url}\fi

\bibitem[Breiman(2001)]{Breiman2001}
Breiman, Leo.
\newblock Random forests.
\newblock \emph{Machine Learning}, 45:\penalty0 5--32, 2001.
\newblock ISSN 0885-6125.
\newblock \doi{10.1023/A:1010933404324}.
\newblock URL \url{http://dx.doi.org/10.1023/A\%3A1010933404324}.

\bibitem[Cai et~al.(2007)Cai, Ee, Pham, Roe, and Zhang]{Cai2007}
Cai, Jinhai, Ee, Dominic, Pham, Binh, Roe, Paul, and Zhang, Jinglan.
\newblock Sensor network for the monitoring of ecosystem: Bird species
  recognition.
\newblock In \emph{Intelligent Sensors, Sensor Networks and Information, 2007.
  ISSNIP 2007. 3rd International Conference on}, pp.\  293--298. IEEE, 2007.

\bibitem[Chawla et~al.(2002)Chawla, Bowyer, Hall, and Kegelmeyer]{ChawlaBHK02}
Chawla, Nitesh~V., Bowyer, Kevin~W., Hall, Lawrence~O., and Kegelmeyer,
  W.~Philip.
\newblock Smote: Synthetic minority over-sampling technique.
\newblock \emph{J. Artif. Intell. Res. (JAIR)}, 16:\penalty0 321--357, 2002.

\bibitem[Eisele et~al.(1996)Eisele, Haeb-Umbach, and Langmann]{Eisele1996}
Eisele, Thomas, Haeb-Umbach, Reinhold, and Langmann, Detlev.
\newblock A comparative study of linear feature transformation techniques for
  automatic speech recognition.
\newblock In \emph{Spoken Language, 1996. ICSLP 96. Proceedings., Fourth
  International Conference on}, volume~1, pp.\  252--255. IEEE, 1996.

\bibitem[Fagerlund(2007)]{Fagerlund2007}
Fagerlund, Seppo.
\newblock Bird species recognition using support vector machines.
\newblock \emph{EURASIP Journal on Advances in Signal Processing}, 2007, 2007.

\bibitem[Kubat \& Matwin(1997)Kubat and Matwin]{KubatM97}
Kubat, Miroslav and Matwin, Stan.
\newblock Addressing the curse of imbalanced training sets: One-sided
  selection.
\newblock In Fisher, Douglas~H. (ed.), \emph{ICML}, pp.\  179--186. Morgan
  Kaufmann, 1997.
\newblock ISBN 1-55860-486-3.

\bibitem[Levy et~al.(2003)Levy, Linares, and Nocera]{Levy2003}
Levy, Christophe, Linares, Georges, and Nocera, Pascal.
\newblock Comparison of several acoustic modeling techniques and decoding
  algorithms for embedded speech recognition systems.
\newblock In \emph{Proceedings of the Workshop on DSP in Mobile and Vehicular
  Systems, Nagoya, Japan}, 2003.

\bibitem[Pazzani et~al.(1994)Pazzani, Merz, Murphy, Ali, Hume, and
  Brunk]{PazzaniMMAHB94}
Pazzani, Michael~J., Merz, Christopher~J., Murphy, Patrick~M., Ali, Kamal,
  Hume, Timothy, and Brunk, Clifford.
\newblock Reducing misclassification costs.
\newblock In \emph{ICML}, pp.\  217--225, 1994.

\bibitem[Quinlan(1993)]{quinlan1993}
Quinlan, J.~Ross.
\newblock \emph{C4.5: Programs for Machine Learning}.
\newblock Morgan Kaufmann, 1993.
\newblock ISBN 1-55860-238-0.

\bibitem[Rabiner \& Wilpon(1979)Rabiner and Wilpon]{Rabiner1979}
Rabiner, L and Wilpon, J.
\newblock Considerations in applying clustering techniques to speaker
  independent word recognition.
\newblock In \emph{Acoustics, Speech, and Signal Processing, IEEE International
  Conference on ICASSP'79.}, volume~4, pp.\  578--581. IEEE, 1979.

\bibitem[Reynolds(1995)]{Reynolds1995}
Reynolds, Douglas~A.
\newblock Speaker identification and verification using gaussian mixture
  speaker models.
\newblock \emph{Speech communication}, 17\penalty0 (1):\penalty0 91--108, 1995.

\bibitem[Rish(2001)]{Rish2001}
Rish, I.
\newblock An empirical study of the naive bayes classifier.
\newblock \emph{IJCAI 2001 Workshop on Empirical Methods in Artificial
  Intelligence}, 3\penalty0 (22):\penalty0 41–46, 2001.
\newblock URL
  \url{http://www.cc.gatech.edu/home/isbell/classes/reading/papers/Rish.pdf}.

\bibitem[Roch\'e et~al.(2009)Roch\'e, Benito-Espinal, and Hautcastel]{CD_MGB}
Roch\'e, Jean~C., Benito-Espinal, Edouard, and Hautcastel, Patricia.
\newblock Oiseaux des antilles.
\newblock Audio CD, 2009.

\bibitem[Rumble et~al.(2001)Rumble, Benkobi, Lindzey, and Gamo]{Rumble2001}
Rumble, M.A., Benkobi, L., Lindzey, F., and Gamo, R.S.
\newblock Evaluating elk habitat interactions with gps collars.
\newblock \emph{Tracking Animals with GPS}, 2001.

\bibitem[Ryan et~al.(2004)Ryan, Petersen, Peters, and Gremillet]{Ryan2004}
Ryan, P.~G., Petersen, S.~L., Peters, G., and Gremillet, D.
\newblock Gps tracking a marine predator: the effects of precision, resolution
  and sampling rate on foraging tracks of african penguins.
\newblock \emph{Marine Biology}, 145(2), 2004.

\bibitem[Sakoe \& Chiba(1978)Sakoe and Chiba]{Sakoe1978}
Sakoe, Hiroaki and Chiba, Seibi.
\newblock Dynamic programming algorithm optimization for spoken word
  recognition.
\newblock \emph{Acoustics, Speech and Signal Processing, IEEE Transactions on},
  26\penalty0 (1):\penalty0 43--49, 1978.

\bibitem[Schedl et~al.(2007)Schedl, Knees, Seyerlehner, and Pohle]{Schedl2007}
Schedl, Markus, Knees, Peter, Seyerlehner, Klaus, and Pohle, Tim.
\newblock The comirva toolkit for visualizing music-related data.
\newblock In \emph{Proceedings of the 9th Joint Eurographics/IEEE VGTC
  conference on Visualization}, pp.\  147--154. Eurographics Association, 2007.

\bibitem[Stattner et~al.(2010)Stattner, Collard, Hunel, and
  Vidot]{Stattner2010-MoMM}
Stattner, Erick, Collard, Martine, Hunel, Philippe, and Vidot, Nicolas.
\newblock Detecting movement patterns with wireless sensor networks:
  application to bird behavior.
\newblock In \emph{MoMM}, pp.\  251--258, 2010.

\bibitem[Stattner et~al.(2011)Stattner, Collard, Hunel, and
  Vidot]{Stattner2011-LCN}
Stattner, Erick, Collard, Martine, Hunel, Philippe, and Vidot, Nicolas.
\newblock Wireless sensor networks for social network data collection.
\newblock In \emph{IEEE Conference on Local Computer Networks (LCN)}, pp.\
  867--874, 2011.

\bibitem[Valin et~al.(2003)Valin, Michaud, Rouat, and
  L{\'e}tourneau]{Valin2003}
Valin, J-M, Michaud, Fran{\c{c}}ois, Rouat, Jean, and L{\'e}tourneau, Dominic.
\newblock Robust sound source localization using a microphone array on a mobile
  robot.
\newblock In \emph{Intelligent Robots and Systems, 2003.(IROS 2003).
  Proceedings. 2003 IEEE/RSJ International Conference on}, volume~2, pp.\
  1228--1233. IEEE, 2003.

\end{thebibliography}

\end{document}